\documentclass{article}

\usepackage[preprint]{corl_2024} %
\usepackage{algorithm}
\usepackage{algpseudocode}
\usepackage{amsmath}
\usepackage{mathtools}
\usepackage[capitalise,nameinlink]{cleveref}
\usepackage{svg}
\usepackage{wrapfig}
\usepackage{multirow}
\usepackage{caption}
\usepackage{subcaption}
\usepackage{booktabs}
\usepackage{array}
\usepackage{tablefootnote}
\usepackage{siunitx}
\usepackage[export]{adjustbox}

\graphicspath{ {./images/} }

\definecolor{darkgreen}{rgb}{0.0, 0.5, 0.0}
\newcommand{\boldgreen}[1]{\textbf{\textcolor{darkgreen}{#1}}}

\newcommand{\greenaug}{\texttt{\boldgreen{GreenAug}}}
\newcommand{\greenrand}{\texttt{\greenaug-Rand}}
\newcommand{\greengen}{\texttt{\greenaug-Gen}}
\newcommand{\greenmask}{\texttt{\greenaug-Mask}}

\title{Green Screen Augmentation Enables Scene Generalisation in Robotic Manipulation}

\author{
  \small{Eugene Teoh$^{1,*}$, Sumit Patidar$^{1,*}$, Xiao Ma$^{1}$, Stephen James$^{1}$} \\
  \small{$^{1}$Dyson Robot Learning Lab, $^{*}$Equal Contribution}\\[0.3cm]
  \normalsize{\textbf{\href{https://greenaug.github.io/}{\texttt{greenaug.github.io}}}}
}

\begin{document}
\maketitle

\begin{abstract}
Generalising vision-based manipulation policies to novel environments remains a challenging area with limited exploration. Current practices involve collecting data in one location, training imitation learning or reinforcement learning policies with this data, and deploying the policy in the same location. However, this approach lacks scalability as it necessitates data collection in multiple locations for each task.
This paper proposes a novel approach where data is collected in a location predominantly featuring green screens. We introduce \boldgreen{Green}-screen \boldgreen{Aug}mentation (\greenaug), employing a chroma key algorithm to overlay background textures onto a green screen. Through extensive real-world empirical studies with over 850 training demonstrations and 8.2k evaluation episodes, we demonstrate that \greenaug{} surpasses no augmentation, standard computer vision augmentation, and prior generative augmentation methods in performance.
While no algorithmic novelties are claimed, our paper advocates for a fundamental shift in data collection practices. We propose that real-world demonstrations in future research should utilise green screens, followed by the application of \greenaug{}. We believe \greenaug{} unlocks policy generalisation to visually distinct novel locations, addressing the current scene generalisation limitations in robot learning.
\end{abstract}


\begin{figure}[h]
    \centering
    \includegraphics[width=0.9\textwidth]{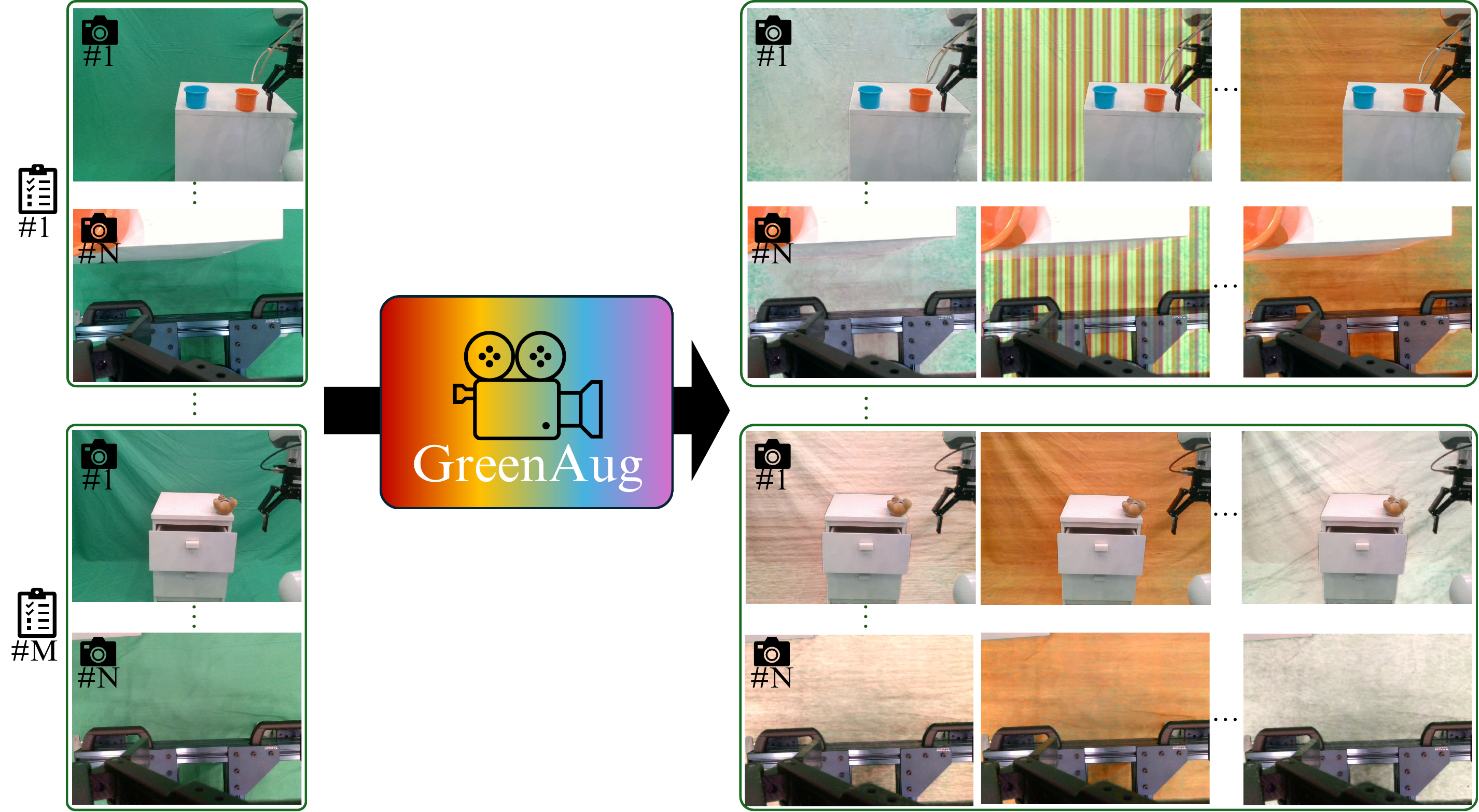}
    \caption{GreenAug provides a simple visual augmentation to robot policies by first collecting data with a green screen, then augmenting it with different textures. The resulting policy can be transferred to unseen visually distinct novel locations (scenes).}
    \label{fig:overview}
\end{figure}

\section{Introduction}

Recent advancements in robot learning policies~\citep{yarats2020drq,yarats2021drqv2,shridhar2023peract,zhao2023act,chi2023diffusionpolicy,ma2024hierarchical,vosylius2024render} have shown significant capabilities in performing complex manipulation tasks. However, generalising these policies to new locations remains a substantial challenge due to the lack of diverse training datasets. Ideally, these datasets should include a wide variety of environments, such as diverse areas of homes. However, gathering real-world data from different scenes is difficult and costly. These scenes refer to visually distinct physical locations, such as an oven situated in different kitchens or a toilet placed in various homes. The difficulty of collecting diverse data necessitates more efficient use of existing datasets.

Generative augmentation approaches~\citep{mandi2022cacti,chen2023genaug,yu2023rosie} have attempted to address this by using generative models~\citep{sohl2015deep,rombach2022stablediffusion,sauer2023adversarial} to augment robot datasets. However, these methods often require extensive manual tuning and face several challenges. This includes text prompt engineering, chaining multiple object detectors, segmenters and generative models, and problems with performance and processing speed. Additionally, they can be inaccurate in robotic settings---particularly in segmentation and inpainting from wrist camera views, potentially introducing noise into robot policies.

In light of these complications, we opt for a simpler yet effective alternative: green screens. The film industry has utilised green screens extensively~\citep{smith1996blue,grundhofer2010color,foster2014green,aksoy2016interactive,smirnov2023magenta}, enabling the addition of virtual backgrounds to live footage. Inspired by these applications, we apply green screen technology to robotics, allowing robots to perform tasks in unfamiliar scenes not part of the training data.

\begin{wrapfigure}[17]{r}{0.5\textwidth}
  \vspace{-2em}
  \begin{center}
    \includegraphics[width=\linewidth]{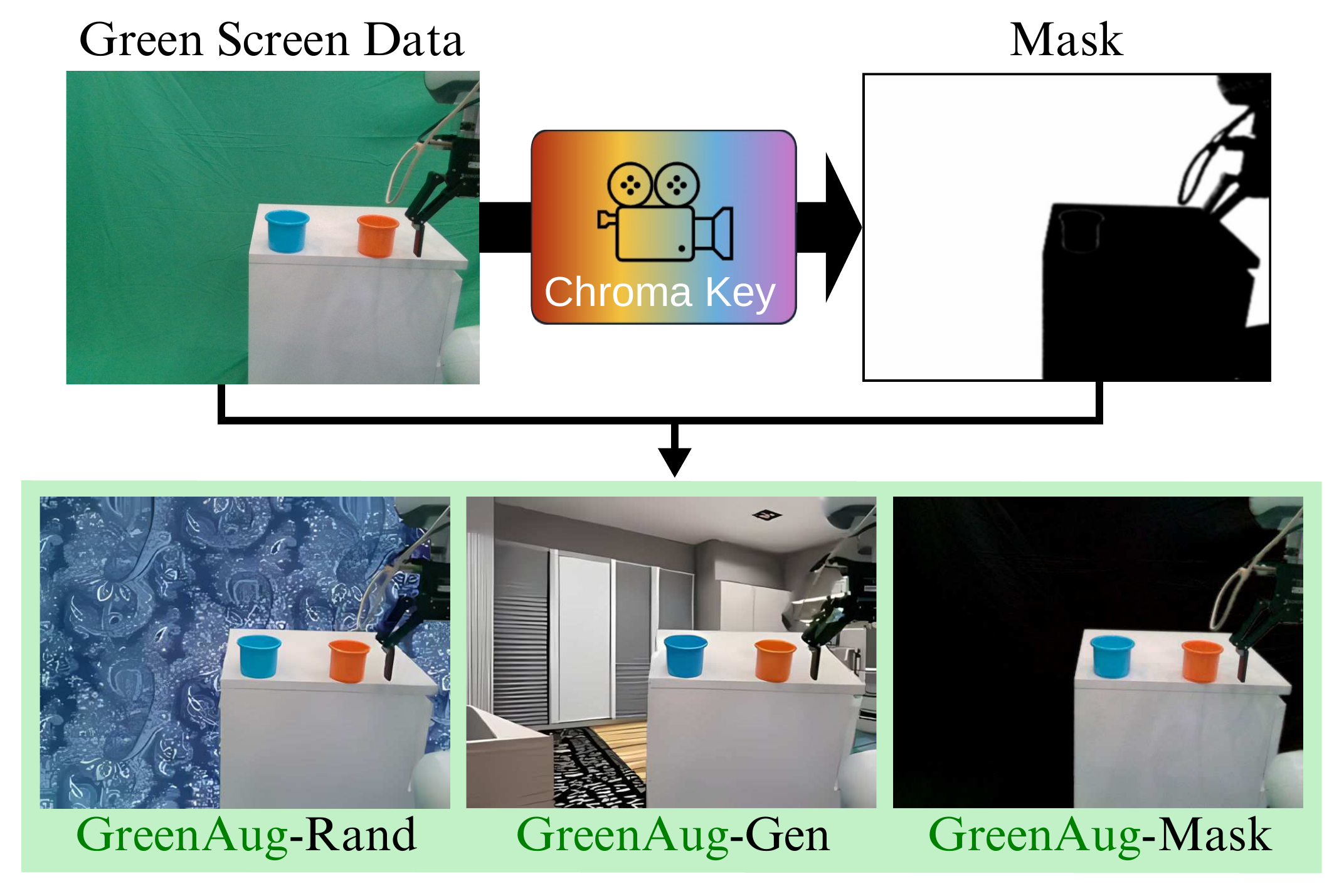}
  \end{center}
  \caption{The \greenaug{} process begins with acquiring a green screen mask using chroma keying. \greenrand{} applies random textures, \greengen{} uses generative models to inpaint the background, and \greenmask{} learns a masking network to filter out the background.}
    \label{fig:method}
\end{wrapfigure}

In this paper, we introduce \boldgreen{Green}-screen \boldgreen{Aug}mentation (\greenaug), a simple real-world visual augmentation method that uses green screen and chroma keying to replace backgrounds, applicable to RGB-based robot learning methods. We explore several variants of \greenaug{}, including the use of random textures (\cref{fig:overview}), backgrounds generated by generative models, and a background masking network to obscure the background during inference. By replacing backgrounds with various textures, it allows robot learning policies to be robust against changes in visual scenes and focus on crucial features in the image space.

We conducted extensive real-world experiments across eight challenging robotic manipulation tasks and six further studies, amounting to over 850 training demonstrations and 8.2k evaluation episodes. We evaluated the performance of control policies in unseen scenes for head-to-head comparisons on scene generalisation. We compared several variants of \greenaug{} against approaches with no augmentation, standard computer vision augmentations, and a generative augmentation~\citep{mandi2022cacti,chen2023genaug,yu2023rosie} method. Our results show that \greenaug{} outperforms no augmentation by 65\%, standard computer vision augmentation by 29\% and generative augmentation by 21\%.

\section{Related Work}

\textbf{Visual augmentation in robotics.} Visual augmentation is important in robotics for adapting to changing environments. Standard computer vision augmentations like random photometric distortion, cropping, shifting, convolutions and overlays have enhanced performance in imitation learning~\citep{young2021visual,xie2023decomposing} and reinforcement learning~\citep{yarats2020drq,laskin2020reinforcement,yarats2021drqv2, hansen2021svea,hansen2021generalization,almuzairee2024sada}. However, most of these methods only apply simple photometric perturbations. Domain randomisation~\citep{Sadeghi-RSS-17, tobin2017domain, james2017transferring, matas2018sim, james2019sim,alghonaim2021benchmarking,so2022sim} enhances this by generating synthetic data with varied visual and physical dynamics parameters for simulation-to-reality (Sim2Real) transfer. Alternatively, methods like CACTI~\citep{mandi2022cacti}, GenAug~\citep{chen2023genaug}, and ROSIE~\citep{yu2023rosie} use generative models such as Stable Diffusion~\citep{rombach2022stablediffusion} to diversify visual data directly on real-world data, bypassing the need for simulation.

\textbf{Green screen in machine learning and robotics.} Green screen has been traditionally used for film and video production~\citep{smith1996blue,grundhofer2010color,foster2014green,aksoy2016interactive,smirnov2023magenta}. In recent years, its application in machine learning has increased. \citet{smirnov2023magenta} applied machine learning to improve the quality of chroma keying. \citet{xu2017deep, sengupta2020background, lin2021real, lin2022robust} explored machine learning techniques to replace green screens, enabling natural image matting without them. \citet{schulein2023comparison} used green screens and chroma keying to replace backgrounds with clinical scenes to create synthetic data for medical clothing detection. In robotics, the use of green screens remains limited. \citet{coates2010multi} used it to develop a multi-camera object detector with synthetic data from chroma-keyed backgrounds.

\section{Green Screen Augmentation} \label{sec:method}

In this section, we provide a detailed introduction to \greenaug{}. The practical steps for \greenaug{} are as follows: \textbf{(1)} Green Screen Scene Setup; \textbf{(2)} \greenaug{} via Chroma Keying; \textbf{(3)} Training Robot Learning Policies. In the following sub-sections, we expand on each of these stages. 

\subsection{Green Screen Scene Setup}

\begin{wrapfigure}[13]{r}{0.6\textwidth}
  \vspace{-3em}
  \begin{center}
    \includegraphics[width=\linewidth]{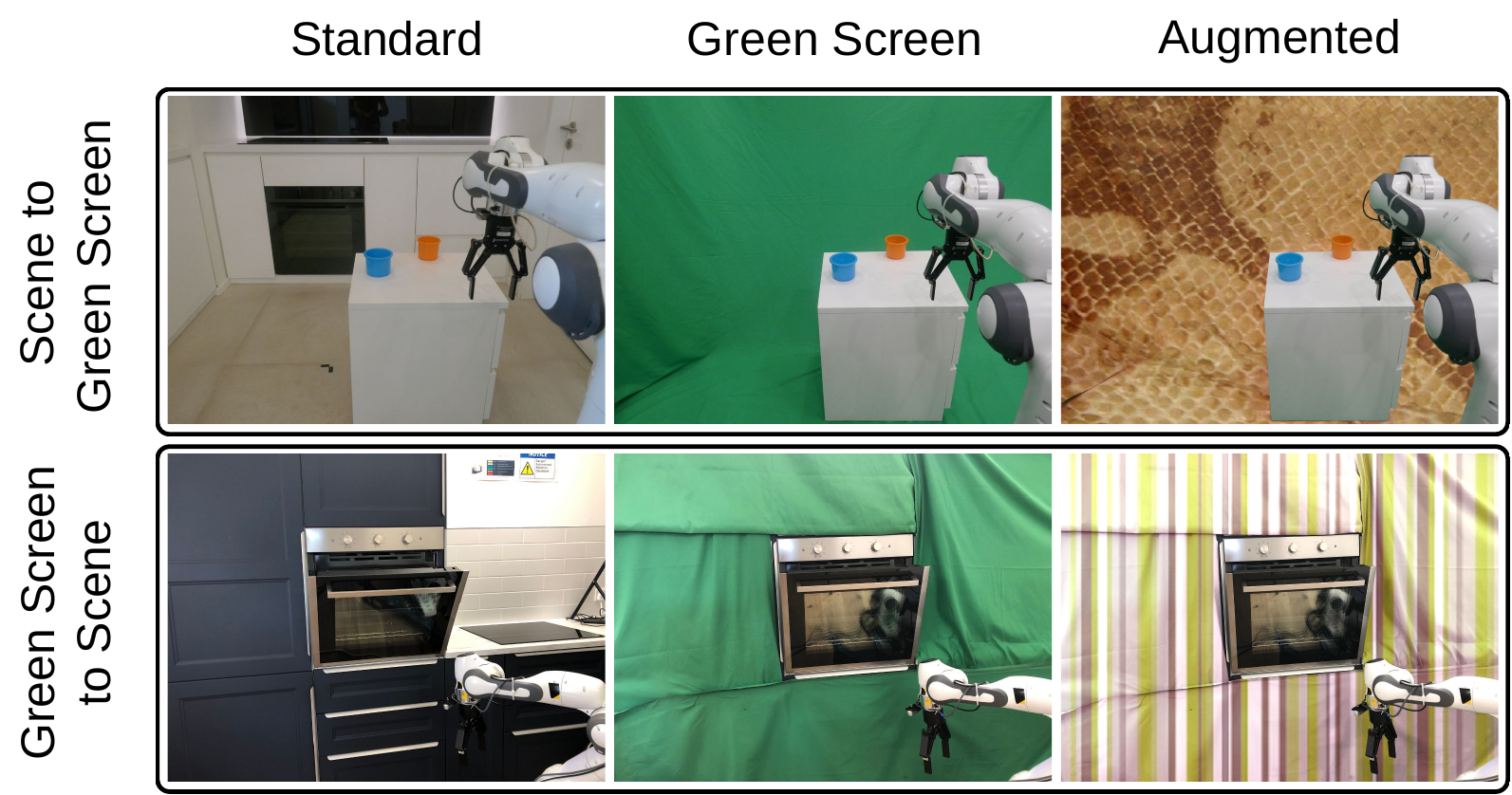}
  \end{center}
  \caption{Physical steps for green-screen setup. Scene items can either be moved into the green screen, or the green screen can be brought to the scene.}
  \label{fig:scene-setup}
\end{wrapfigure}

The act of scene setup consists of obscuring the background (i.e. non-task relevant objects) with a green screen. There are several ways of achieving this, two of which are highlighted in \cref{fig:scene-setup} and described below. Once the scene has been set up, demonstration collection can begin.

\textbf{Scene to Green Screen}, where a permanent green screen area or room is established, and items can be moved into the green screen for data collection. This is the most common use case and includes tasks such as general pick-and-place, opening drawers, sweeping, pushing, etc. 

\textbf{Green Screen to Scene}, where the green screen is brought to a fixed, unmovable object. Scenes that usually fall into this category are ones that require manipulating integrated or heavy objects, such as stacking dishwashers, opening ovens, and opening doors.

\subsection{GreenAug via Chroma Keying}

Chroma keying is a visual effects technique for layering two images or video streams together based on colour hues (chroma range). This technique is commonly used in video production and post-production to composite two frames or images together by removing a background colour (usually green or blue) from the foreground content, making it transparent. This allows for the insertion of a new background or visual element in place of the green or blue background. Many chroma key algorithms exist, but we opt for a simple algorithm proposed by \citet{chromakey}.
Given the generated mask, several options are available for applying \greenaug{}. We provide three variants of \greenaug{}: Random (\greenrand{}), Generative (\greengen{}) and Mask (\greenmask{}), illustrated in \cref{fig:method} and described in detail below.

\textbf{\greenrand{}} This variant applies a fixed set of random textures to the chroma-keyed background. Following research in domain randomisation~\citep{Sadeghi-RSS-17, tobin2017domain, james2017transferring, matas2018sim, james2019sim,alghonaim2021benchmarking,so2022sim}, increasing the variability of these textures helps the policy ignore the background and focus on task-specific items (objects manipulated by the policy).

\textbf{\greengen{}.} This variant uses the chroma-keyed mask to inpaint realistic or imagined backgrounds using generative models like Stable Diffusion. Examples of prompts include: ``photorealistic bedroom'', ``photorealistic kitchen'', ``photorealistic living room''. This method augments the image with semantic backgrounds, aiming to closely resemble real-world scenarios.

\textbf{\greenmask{}.} This variant uses a masking (soft segmentation) network trained to predict masks. These predicted masks are then applied to the image observations to obtain blacked-out, dark backgrounds. This simplification of the visual input potentially helps the visuomotor policies to focus on the main elements of interest by eliminating background noise and distractions. During training, the masking network processes images against chroma-keyed backgrounds with random textures (akin to \greenrand{}) and learns to predict the masks generated through chroma keying.

\begin{table}[t]
\centering
\caption{Main experiment results averaged across three novel scenes. Each task-method combination is evaluated with 112 evaluation episodes on average. Full detailed results are provided in the Appendix.}
\label{tab:results}

\resizebox{\textwidth}{!}{%
\begin{tabular}{@{}ccccccc@{}}
\toprule
 & \multicolumn{6}{c}{Success Rate (\%)} \\ \cmidrule(l){2-7} 
Task & NoAug & CVAug & \begin{tabular}[c]{@{}c@{}}Generative\\ Augmentation\end{tabular} & \begin{tabular}[c]{@{}c@{}}\greenaug{}\\ \texttt{Rand}\end{tabular} & \begin{tabular}[c]{@{}c@{}}\greenaug{}\\ \texttt{Gen}\end{tabular} & \begin{tabular}[c]{@{}c@{}}\greenaug{}\\ \texttt{Mask}\end{tabular} \\ \midrule
Open Drawer & 59 & 65 & 77 & \textbf{96} & 87 & 79 \\
Place Cube in Drawer & 36 & 69 & 70 & \textbf{92} & 83 & 37 \\
Take Lid off Saucepan & 67 & 81 & 77 & \textbf{88} & 73 & 71 \\
Sweep Coffee Beans & 66 & 78 & 75 & \textbf{96} & 81 & 84 \\
Place Jeans in Basket & 71 & 75 & 76 & \textbf{87} & 77 & 67 \\
Place Bear in Basket & 45 & 63 & 61 & \textbf{95} & 49 & 41 \\
Stack Cups & 49 & 59 & 77 & \textbf{81} & 72 & 55 \\
Slide Book and Pick Up & 49 & 74 & 89 & \textbf{93} & \textbf{93} & 35 \\ \midrule
Average & 55 & 70 & 75 & \textbf{91} & 77 & 58 \\ \bottomrule
\end{tabular}%
}
\vspace{-0.5cm}
\end{table}

\subsection{Training Robot Learning Policies}

\greenaug{} can be applied to RGB-based robot learning methods. Similar to standard augmentation methods, images can be transformed with \greenaug{} and fed into policy networks during training, or they can be preprocessed offline and then used for training. Offline preprocessing is more common due to the longer computation time of some \greenaug{} variants. However, in online settings such as reinforcement learning, online transformations are also effective. \greenrand{} and \greengen{} allow each raw frame from the training demonstrations to be augmented with different textures, significantly increasing the amount of preprocessed data. In contrast, \greenmask{} masks the background and only provides a one-to-one mapping of the original to augmented trajectory. To ensure a fair comparison, we keep the number of preprocessed frames equal to the number of raw frames for all methods.

In our main experiment (\cref{sec:results}), we chose Action Chunking with Transformers (ACT)~\citep{zhao2023act} as our control variable to demonstrate the effectiveness of this augmentation method. We selected ACT because of its recent success in adapting behaviours from a modest number of demonstrations, making it an ideal platform to showcase the benefits of \greenaug{}. Additionally, in \cref{sec:further-studies}, we demonstrate that \greenrand{} is also effective with a reinforcement learning policy.

\section{Experiments}

In this section, we present the experiments to evaluate the effectiveness of \greenaug{} on robot learning policies. Prior works~\citep{xie2023decomposing, pumacay2024colosseum} have confirmed the effectiveness of background and texture randomisation in simulation. Since \greenaug{} focuses on real-world data augmentation, our experiments are conducted exclusively in the real world. We aim to study the following: \textbf{(1)} Does \greenaug{} improve visual generalization to unseen scenes? \textbf{(2)} Which variant of \greenaug{} is the most effective, and what are the tradeoffs? \textbf{(3)} Is \greenaug{} applicable in different data collection settings? \textbf{(4)} Is \greenaug{} agnostic to robot embodiments and learning methods?

\subsection{Baselines}

We implement several baselines to compare with \greenaug{}, as described below.

\textbf{No augmentation (NoAug).} No visual augmentation.

\textbf{Computer Vision augmentation (CVAug)}. Random photometric distortions and random shift.

\textbf{Generative augmentation.} Generative augmentation encompasses a broader range of methods such as CACTI~\citep{mandi2022cacti}, GenAug~\citep{chen2023genaug}, and ROSIE~\citep{yu2023rosie}.  CACTI uses Stable Diffusion for inpainting but does not detail the method for obtaining object masks. GenAug, on the other hand, is constrained to a tabletop setting. ROSIE relies on proprietary models and does provide publicly available code. Thus, we have developed our own implementation that closely aligns with these methods. Our implementation is based on Grounding DINO~\citep{liu2023grounding} for open vocabulary object detection, Segment Anything~\citep{kirillov2023segment} for zero-shot segmentation, and Stable Diffusion Turbo~\citep{rombach2022stablediffusion, sauer2023adversarial} for inpainting, integrated with ControlNet~\citep{zhang2023controlnet} and conditioned on DPT-Hybrid~\citep{ranftl2021dpt} (monocular depth estimator) for better generation. Generative augmentation is similar to \greengen{}, but it uses object detection and segmentation for mask creation instead of chroma keying. The pseudocode detailing this implementation is outlined in the Appendix.

\subsection{Setup}

\begin{wrapfigure}[11]{r}{0.6\textwidth}
    \vspace{-4em}
    \begin{center}
        \includegraphics[width=\linewidth]{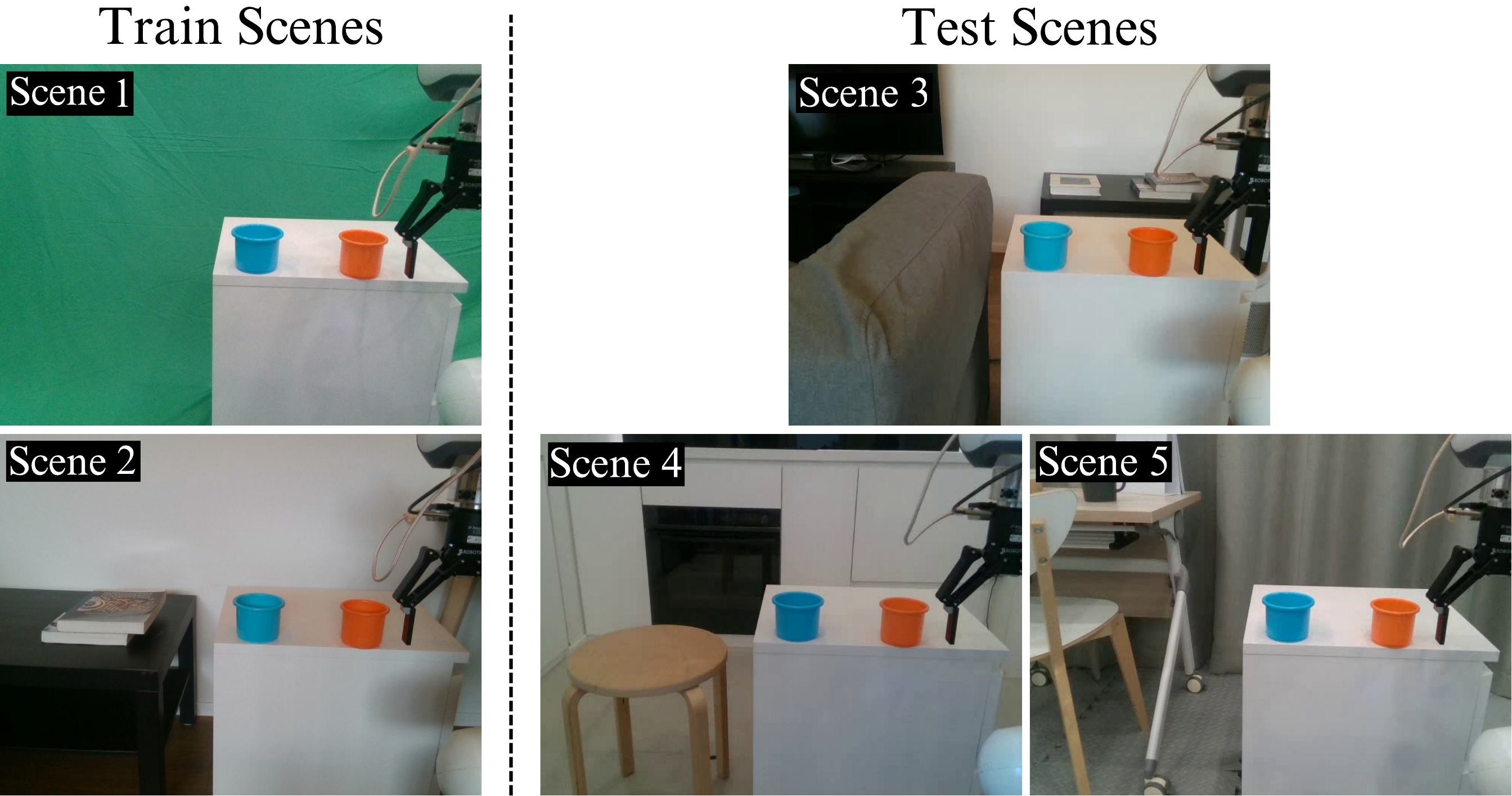}
    \end{center}
    \caption{Visualisations of train and test scenes.}
    \label{fig:train-test-scenes}
\end{wrapfigure}

\begin{figure}[t]
    \centering
    \includegraphics[width=\textwidth]{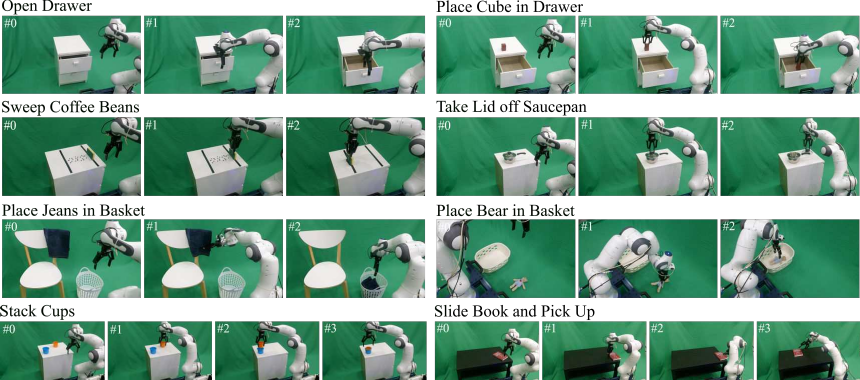}
     \caption{Visualisations of real-world tasks. The trajectory sequences are stacked horizontally, starting with the initial positions labelled as \#0.}
    \label{fig:tasks}
\end{figure}

For our main experiment, we designed eight tasks (illustrated in \cref{fig:tasks}) and structured our experiments for each task as follows.

\textbf{Data collection.} We collected two sets of demonstrations, each consisting of 50 demos. One set was recorded against a green screen (Scene 1), and the other within a standard setting (Scene 2). All data were collected using a leader-follower teleoperation system, similar to ALOHA~\citep{zhao2023act}, but with a 7-DoF Franka Panda arms and a 2F-140 Robotiq gripper on the follower. We used three D415 Realsense cameras, positioned at the upper wrist, lower wrist and left shoulder camera. The images are captured at a resolution of 240 (height) x 320 (width) pixels. For the main experiments alone, we collected over 800 demonstrations and conducted more than 6.6k evaluation runs. Additionally, we gathered about 50 more training demonstrations and 1.6k evaluations for the ablation and further studies in \cref{sec:further-studies}.

\textbf{Training.} We trained all baselines and our methods on both sets of data, except for \greenaug{}, which was excluded from Scene 2 as it relies on the green screen. Each data set corresponds to a separate policy. ACT is used as the control policy for our main experiments. We use absolute joint positions as actions.

\textbf{Evaluation.} In addition to Scenes 1 and 2, we evaluated the methods in three novel scenes (Scenes 3--5). Initially, each method was assessed in Scene 1 to establish an upper-bound performance for the task. Subsequently, the methods were evaluated in Scene 3--5 to test generalisation. For each combination of task, method, train scenes (2), test scenes (3), we performed 25 evaluation runs.

Each scene is shown in \cref{fig:train-test-scenes}. To focus on testing visual generalisation across different scenes, we maintained the positions and orientations of the objects (while applying the same degree of randomisation for one-to-one comparison) relative to the robot while moving between scenes.

\begin{table}[t]
    \centering
    \begin{minipage}[t]{0.49\textwidth}
        \centering
        \caption{Processing time per RGB frame.}
        \label{tab:speed-memory}
        \adjustbox{valign=t}{%
            \begin{tabular}{@{}ll@{}}
            \toprule
            Method & Time ($\downarrow$) \\ \midrule
            Generative Augmentation & 2.530 s \\
            \greenrand{} & \textbf{0.009 s} \\
            \greengen{} & 0.882 s \\ \bottomrule
            \end{tabular}
        }
    \end{minipage}
    \hfill
    \begin{minipage}[t]{0.49\textwidth}
        \centering
        \caption{\greenrand{} applied to RL.}
        \label{tab:rl}
        \adjustbox{valign=t}{%
            \begin{tabular}{@{}cccc@{}}
            \toprule
             &  & \multicolumn{2}{c}{Success Rate (\%)} \\ \cmidrule(l){3-4} 
            \begin{tabular}[c]{@{}c@{}}Train\\ Scene\end{tabular} & \begin{tabular}[c]{@{}c@{}}Test\\ Scene\end{tabular} & NoAug & \begin{tabular}[c]{@{}c@{}}\greenaug\\ \texttt{Rand}\end{tabular} \\ \midrule
            \begin{tabular}[c]{@{}c@{}}Green\\ Screen\end{tabular} & \begin{tabular}[c]{@{}c@{}}1 Novel\\ Scene\end{tabular} & 12 & \textbf{64} \\ \bottomrule
            \end{tabular}%
        }
    \end{minipage}

    \begin{minipage}[t]{0.49\textwidth}
        \centering
        \caption{\greenrand{} with different texture types averaged across tasks and novel scenes. Entropy signifies the amount of texture randomness.}
        \label{tab:texture-randomness}
        \adjustbox{valign=t}{%
            \begin{tabular}{@{}ccc@{}}
            \toprule
            Texture Type & \begin{tabular}[c]{@{}c@{}}Entropy\\ (bits)\end{tabular} & \begin{tabular}[c]{@{}c@{}}Success\\ Rate (\%)\end{tabular} \\ \midrule
            None & - & 48 \\
            Solid Colours & 0.00 & 65 \\
            Perlin Noise & 4.45 & 66 \\
            MIL Textures & 6.81 & \textbf{87} \\ \bottomrule
            \end{tabular}
        }
    \end{minipage}
    \hfill
    \begin{minipage}[t]{0.49\textwidth}
        \centering
        \caption{Object generalisation results. Policies trained on a green cup were tested on other objects. (n) specifies the number of objects tested in the category.}
        \label{tab:object-generalisation}
        \adjustbox{valign=t}{%
            \begin{tabular}{@{}cccc@{}}
            \toprule
             & \multicolumn{3}{c}{Success Rate (\%)} \\ \cmidrule(l){2-4} 
            \begin{tabular}[c]{@{}c@{}}Object\\ Category\end{tabular} & NoAug & \begin{tabular}[c]{@{}c@{}}\greenaug{}\\ \texttt{Rand}\end{tabular} & \begin{tabular}[c]{@{}c@{}}\greenaug{}\\ \texttt{Gen}\end{tabular} \\ \midrule
            Cups (3) & \textbf{95} & 83 & 80 \\
            Cans (2) & 38 & \textbf{46} & 40 \\
            Cubes (2) & 0 & 32 & \textbf{52} \\
            Soft Toy (1) & 0 & \textbf{84} & 72 \\ \midrule
            Average & 45 & 61 & \textbf{62} \\ \bottomrule
            \end{tabular}
        }
    \end{minipage}
    \vspace{-2em}
\end{table}

\subsection{Results}\label{sec:results}

\begin{wrapfigure}[17]{R}{0.65\textwidth}
    \centering
    \includegraphics[width=\linewidth]{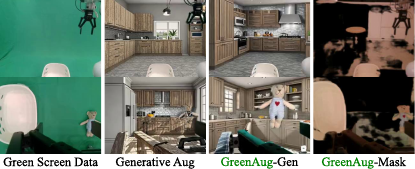}
    \caption{Visualisations of raw and preprocessed frames (left shoulder and lower wrist camera views) of generative augmentation, \greengen{} and \greenmask{} (during inference). Both generative methods struggle with producing good contextual wrist camera inpainting. In generative augmentation, the gripper is inpainted as part of the background, while \greenmask{} shows masking artefacts in novel scenes.}
    \label{fig:aug-noise}
\end{wrapfigure}

\cref{tab:results} presents our experimental findings. The results demonstrate that \greenrand{} surpasses all other baseline methods across all tasks. Specifically, \greenrand{} shows approximately a 65\% improvement over NoAug, around a 29\% improvement compared to CVAug, and about a 21\% improvement over generative augmentation.

Surprisingly, \greengen{} and generative augmentation rank second and third in performance respectively, despite using semantically meaningful backgrounds like living rooms or kitchens. As expected, both methods perform similarly, since they differ only in how they obtain background masks (object detection and segmentation). This suggests that specific semantic content is not crucial for \greenaug{}'s success, as the variant using random backgrounds performs even better. This superior performance may have resulted from the wider variety of colours and textures offered by the random backgrounds.

Generative augmentation performs slightly worse than \greengen{}, likely because it struggles to provide good masks in wrist camera views (illustrated in \cref{fig:aug-noise}), which are essential for tasks requiring precise and stable visual input. Despite advancements in generative models, segmentation and inpainting from robot camera views remain suboptimal.

\greenmask{} shows the least effectiveness among all methods tested. Qualitative evaluations of the masked images reveal frequent failures to completely obscure backgrounds, especially in novel scenes (shown in \cref{fig:aug-noise}). This issue stems from two main factors: the inherent imperfections in ground truth masks obtained from chroma keying and the compounding error from the masking network. The network's imperfect masking further complicates the tasks, pushing the images into out-of-distribution states that challenge the control policy.

\subsection{Ablation and Further Studies}\label{sec:further-studies}

\begin{figure}[t]
    \centering
    \begin{subfigure}[t]{0.49\textwidth}
        \centering
        \includegraphics[width=\linewidth]{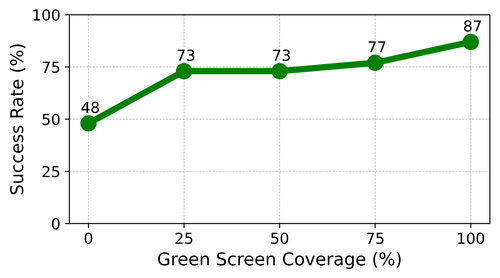}
        \caption{\label{fig:green_screen_coverage}}
    \end{subfigure}
    \hfill
    \begin{subfigure}[t]{0.49\textwidth}
        \centering
        \includegraphics[width=0.7\linewidth]{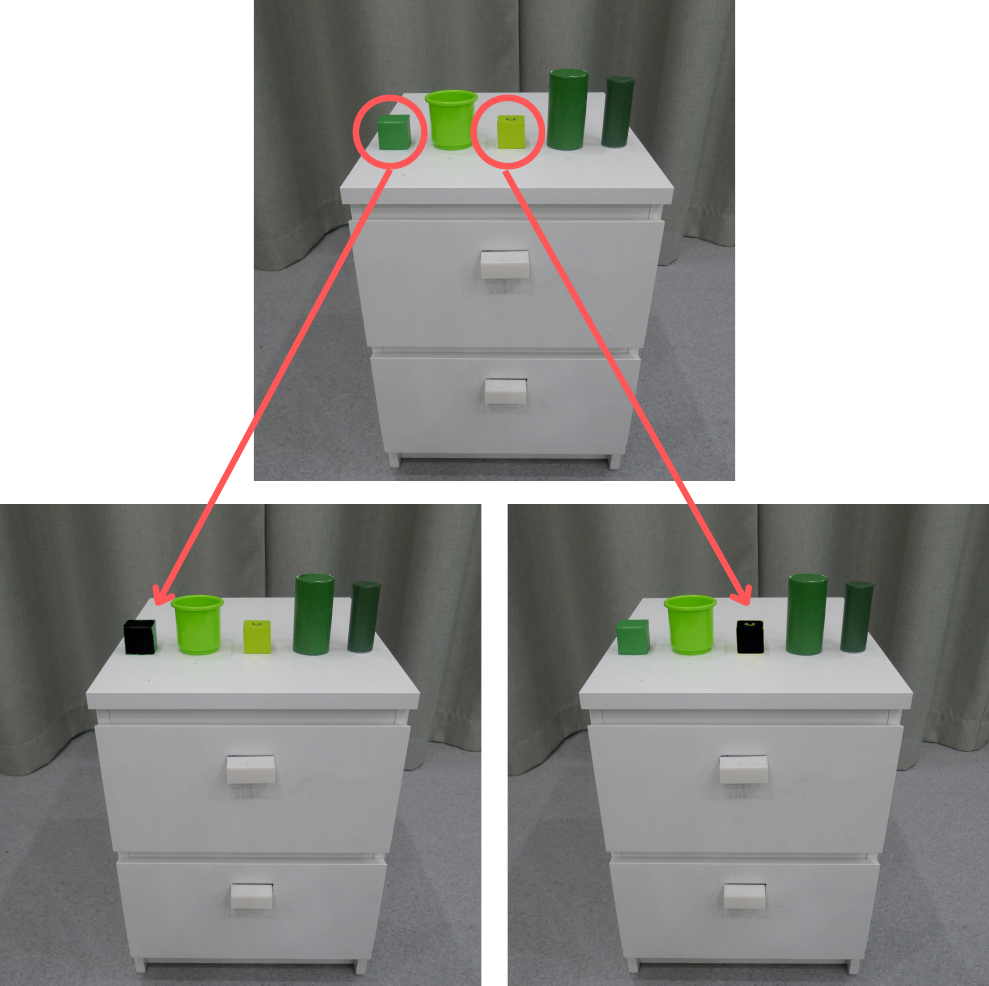}
        \caption{\label{fig:multi-green}}
    \end{subfigure}
    \caption{(\subref*{fig:green_screen_coverage}) \greenrand{} performs the best when applied to all frames per trajectory. (\subref*{fig:multi-green}) Visual assessment of applying \greenaug{} to a single shade of green, in scenarios where multiple objects with varying shades of green are present. Masked objects have their backgrounds blacked out.}
    \label{fig:combined-figures}
    \vspace{-1em}
\end{figure}

Based on the main experiments, we demonstrated that \greenrand{} outperforms all other methods. We then conducted the following in-depth analyses.

\textbf{Benchmarking GreenAug's speed.} We conducted a benchmark to compare the processing speed of various methods, shown in \cref{tab:speed-memory}. CVAug and \greenmask{} were excluded because the former is applied on the fly during training, and the latter performs poorly. We show that \greenrand{} is significantly faster than the other two generative methods.

\textbf{Applying GreenAug to a different robot with reinforcement learning.} We investigated whether \greenaug{} can be applied to a different robot embodiment and learning method, beyond the Franka Panda and ACT. We set up a similar ``take lid off saucepan" task on a UR5. We used Coarse-to-fine Q-network~\citep{seocontinuous} for the policy. The robot was provided with 24 demonstrations and was given a sparse reward of 0 for failure and 1 for success. We trained the robot online with 20 minutes of exploration on a green screen background and evaluated two policies, NoAug and \greenrand{}, in one novel scene. The results, shown in \cref{tab:rl} demonstrate that \greenrand{} applied to reinforcement learning with a different robot performs significantly better than NoAug.

\textbf{Impact of texture randomness.} We investigated how the texture randomness of \greenrand{} affects performance. We tested solid colours, Perlin noise (procedurally generated textures)~\citep{perlin1985perlin}, and MIL textures~\citep{finn2017one} (used in the main experiments). All texture datasets are of the same size (5771). The evaluation was conducted on the ``put cube in drawer'' and ``stack cups'' tasks from the main experiment across three novel scenes (Scenes 3--5). \cref{tab:texture-randomness} summarises the results. Consistent with domain randomization studies~\citep{Sadeghi-RSS-17, tobin2017domain, james2017transferring, matas2018sim, james2019sim, so2022sim}, greater texture randomness leads to better performance. Examples of each texture type are provided in the Appendix.

\textbf{Generalisation across object category.} We assessed if \greenaug{} can be applied not just to backgrounds but also to different object categories. We set up a simple pick-and-place task. We first trained on a green cup and then tested on other visually different objects. The results, shown in \cref{tab:object-generalisation}, indicate that \greengen{} performs best, with only a 1\% difference from \greenrand{}. Both methods outperform NoAug by more than 35\%. NoAug performs well on cups but fails with cubes and soft toys, and occasionally works with cans due to their similar geometric shapes to cups. \greenrand{} and \greengen{} show better performance across different object categories, demonstrating some level of generalisation. However, performance with cups suffers slightly, likely due to the strong augmentation causing confusion about geometric shapes.

\textbf{Green screen coverage.} In real-world settings, some frames in the robot data may move away from the green screen during robot servoing. For example, if the green screen is only partially set up in the scene, the robot may observe parts of the scene not covered by the green screen. To emulate this scenario, we applied \greenrand{} to varying percentages of frames per episode. This was evaluated on the same tasks as the texture randomness study. The results are summarised in \cref{fig:green_screen_coverage}. As expected, green screen coverage is proportional to the success rate.

\textbf{Presence of multiple green objects.} Green screens could affect scenes when there are multiple green objects. We evaluated the sensitivity of chroma keying under these conditions, a challenge also encountered in the film industry. This study questions whether chroma keying can effectively isolate one green object without impacting others. We conducted a visual assessment (shown in \cref{fig:multi-green}) and showed that we can augment only one object at a time while leaving the others unchanged. Alternatively, one can also use a different colour such as blue (along with a green background) for chroma-keying objects.

\section{Conclusion and Limitations}

This paper proposes and investigates the efficacy of \greenaug{} in robotic manipulation across a variety of real-world scenarios. We have demonstrated that \greenaug{} not only works effectively across different tasks but also surpasses other augmentation methods in performance while maintaining simplicity. \greenaug{} outperforms NoAug by approximately 65\%, CVAug by 29\% and generative augmentation by about 21\%. Our findings advocate for a paradigm shift in data collection practices for robot learning. We propose the use of green screens for future real-world demonstrations. Implementing \greenaug{} could significantly improve policy generalisation across novel locations, effectively addressing scene generalisation limitations currently faced in the field.

While \greenaug{} proves to be useful, several challenges remain that we have outlined for future research. \greenaug{} is effective for background generalisation and to an extent, object generalisation (as shown in further studies), but it falls short when it comes to adapting to objects with very different geometric shapes. This type of generalisation involves changing the dynamics and trajectories of the demonstrations, such as accommodating different mugs with unique handles that require specific grasping points. Furthermore, \greenaug{} could be complementary to generative augmentation. This combination could help train world models capable of producing imaginary trajectories that generalise across diverse objects and appliances.

\clearpage

\acknowledgments{Big thanks to the members of the Dyson Robot Learning Lab for discussions and infrastructure help: Nic Backshall, Iain Haughton, Younggyo Seo, Sridhar Sola, Jafar Uruc, Yunfan Lu, Abdi Abdinur, Nikita Chernyadev.}

\bibliography{main}  %

\clearpage

\appendix

\section{Experiment Setups}\label{app:experiment-plans}
In this section, we provide the detailed setups of our real-robot experiments to help reproduce the results.

\begin{figure}[htp]
    \centering
    \begin{subfigure}[t]{0.49\textwidth}
        \centering
        \includegraphics[width=\textwidth]{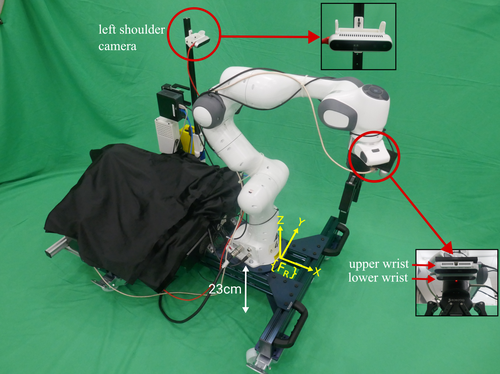}
        \caption{Robot setup. Franka Panda Arm is mounted on a Vention base with three Realsense cameras. The robot is above the ground by \qty{23}{\cm}. The robot pose is represented in the base frame, ${F_R}$.}
        \label{fig:robot_setup}
    \end{subfigure}
    \hfill
    \begin{subfigure}[t]{0.49\textwidth}
        \centering
        \includegraphics[width=\textwidth]{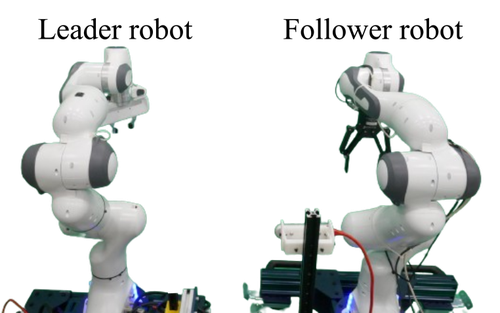}
        \caption{Leader and follower robot setup.}
        \label{fig:leader_follower}
    \end{subfigure}
\end{figure}

\textbf{Robot Setup.} The robot setup consists of a 7-DoF Franka Panda Emika arm equipped with a Robotiq 2F-140 gripper. We use three RealSense D415 cameras: two cameras mounted on the end-effector (lower wrist, upper wrist) for a wide field-of-view, and one camera (left shoulder) fixed on the base, as depicted in Fig. \ref{fig:robot_setup}.

\textbf{Data collection.} We gather demonstrations for our tasks utilising a leader-follower setup similar to ALOHA~\cite{zhao2023act}. An expert human demonstrator moves the Leader arm, and the Follower arm mirrors the Leader's joint positions, as shown in \cref{fig:leader_follower}. Camera and robot state observations are recorded at 30 FPS.

\textbf{Tasks.} For each task, we collect 50 demonstrations each at two scenes: green screen room and living room. \cref{fig:task_schematics} shows the task definitions with sketches to illustrate the setup with measurements and randomisation. For all tasks, the initial robot joint positions are [0.0, -0.785, 0.0, -2.356, 0.0, 1.571, 0.0].

\newpage

\begin{figure}[H]
\begin{subfigure}{\textwidth}
  \centering
\includegraphics[width=\textwidth]{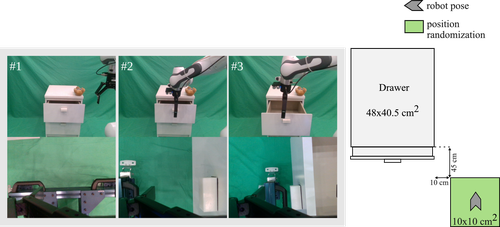}
  \caption{\textbf{Open Drawer:} The robot base ($F_R$) is uniformly randomised inside the $\qty{10}{\cm} \times \qty{10}{\cm}$ region. The gripper then slides into the small drawer opening and then pulls the drawer open. In total, 50 demonstrations are collected with an average demo length of 169 steps or 13 secs.}
  \label{fig:open_drawer_schematics}
\end{subfigure}
\begin{subfigure}{\textwidth}
  \centering
\includegraphics[width=\textwidth]{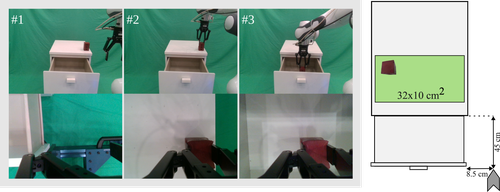}
  \caption{\textbf{Place Cube in Drawer:} The robot base ($F_R$) is fixed relative to the drawer. The cube is randomised on the drawer top within the $\qty{32}{\cm} \times \qty{10}{\cm}$ region. The robot picks up the cube and places it inside the opened drawer. A total of 50 demonstrations are collected with an average demo length of 250 steps or 19 secs.}
  \label{fig:put_cube_schematics}
\end{subfigure}
\begin{subfigure}{\textwidth}
  \centering
\includegraphics[width=\textwidth]{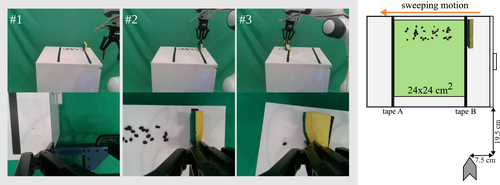}
  \caption{\textbf{Sweep Coffee Beans:} The robot base ($F_R$) is fixed relative to the drawer. The drawer used is the same as in previous tasks but rotated by \qty{90}{\degree}. We stick two black tapes on the drawer top. The coffee beans are randomised in the $\qty{24}{\cm} \times \qty{24}{\cm}$ region between the two tapes. The sponge position is randomised along the tape B (\qty{24}{\cm}). The robot grasps the sponge and sweeps the coffee beans to the left of tape A. A total of 50 demonstrations are collected with an average demo length of 314 steps or 24 secs. For evaluations, a trial is considered successful if 90\% of beans are swept. We use 20 beans, so at least 18 beans needs to be swept for success.}
  \label{fig:sweep_coffee_beans_schematics}
\end{subfigure}
\end{figure}

\begin{figure}\ContinuedFloat
\begin{subfigure}{\textwidth}
  \centering
\includegraphics[width=\textwidth]{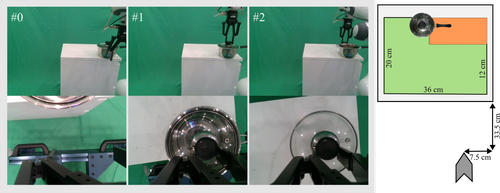}
  \caption{\textbf{Take Lid off Saucepan:} The robot base ($F_R$) is fixed relative to the drawer. The saucepan is randomized in the L-shaped region. The robot grasps the lid of the saucepan and always places it in the orange area. In total, 50 demonstrations were collected with an average demo length of 857 steps.}
  \label{fig:saucepan_schematics}
\end{subfigure}
\begin{subfigure}{\textwidth}
  \centering
\includegraphics[width=\textwidth]{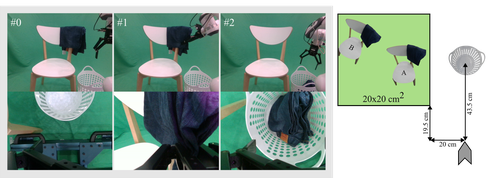}
  \caption{\textbf{Place Jeans to Basket:} The robot base ($F_R$) is fixed relative to the laundry basket. The jeans is semi-folded and hanging on the right edge of the chair. The chair position is randomised in the $\qty{20}{\cm} \times \qty{20}{\cm}$ region. The robot grasps the jeans from the side and places it in the laundry basket. We collected 50 demonstrations, in each half of the demonstrations the chair position is randomised keeping the orientation A and in the other half the orientation of the chair remained B. The average length of the demo is 592 steps or 44 secs.}
  \label{fig:put_jeans_schematics}
\end{subfigure}

\begin{subfigure}{\textwidth}
  \centering
\includegraphics[width=\textwidth]{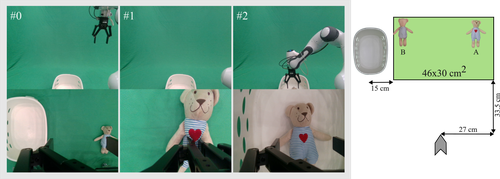}
  \caption{\textbf{Place Bear in Basket:} The robot base ($F_R$) is fixed and the bear (toy) position is randomised in the $\qty{46}{\cm} \times \qty{30}{\cm}$ region on the ground. For half of the demos, the randomisation is done in orientation A and for the other half in orientation B. The robot first picks up the toy and places it in the basket nearby. We collect 50 demonstrations in total with an average demo length of 741 steps or 56 secs.}
\end{subfigure}
\end{figure}

\begin{figure}\ContinuedFloat
\begin{subfigure}{\textwidth}
  \centering
\includegraphics[width=\textwidth]{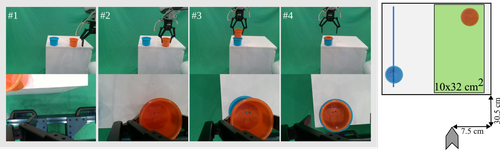}
  \caption{\textbf{Stack Cups:} The robot base ($F_R$) is fixed relative to the drawer. The orange cup is randomised in the $\qty{10}{\cm} \times \qty{32}{\cm}$ region whereas the blue cup is randomised along the blue line (32cm). The robot first picks up the orange cup and stacks it on the blue cup. In total, 50 demonstrations are collected with an average demo length of 590 steps or 44 secs.}
  \label{fig:stack_cups_schematics}
\end{subfigure}

\begin{subfigure}{\textwidth}
  \centering
\includegraphics[width=\textwidth]{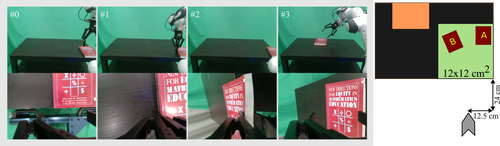}
  \caption{\textbf{Slide Book and Pick Up:} The robot base ($F_R$) is fixed relative to the black coffee table. The book position is randomised in the $\qty{12}{cm} \times \qty{12}{cm}$ region. In half of the demonstrations, the book orientation is kept A and in the other half, orientation B. The robot first corrects the book orientation if necessary by pushing on its edge and then slides the book to the edge of the table. It then picks it up and places it in the area depicted by orange (rightmost figure). We collect 50 demonstrations in total with an average demo length of 930 steps or 70 secs.}
  \label{fig:slide_book_schematics}
\end{subfigure}

\begin{subfigure}{\textwidth}
  \centering
\includegraphics[width=\textwidth]{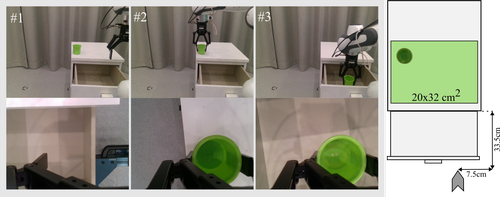}
  \caption{\textbf{Place Cup in Drawer (Object Generalisation):} The robot base ($F_R$) is fixed and the green cup position is randomised in the $\qty{20}{cm} \times \qty{32}{cm}$ region on the drawer top. The robot first picks up the cup and places it in the drawer. We collect 50 demonstrations in total with an average demo length of 597 steps or 45 secs.}
\end{subfigure}
\caption{\textbf{Task definitions with randomisation}. We illustrate 8 main tasks and 1 ablation task with randomisation used. We used the images from the left shoulder (top row) and lower wrist camera (bottom row) to describe each task sequence. Note that the sketches on the right are not drawn to scale.}
\label{fig:task_schematics}
\end{figure}

\begin{figure}[H]
    \centering
    \includegraphics[width=\textwidth]{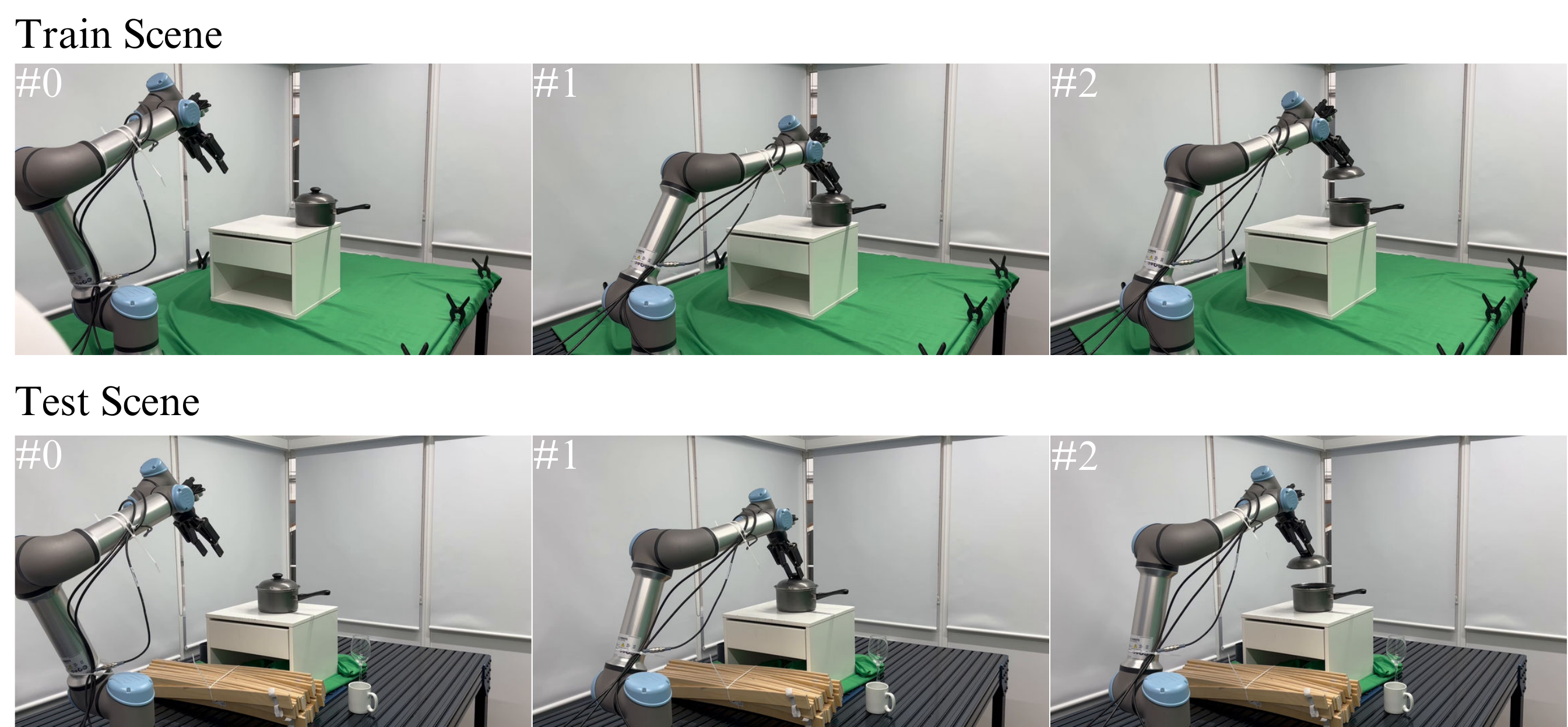}
    \caption{Reinforcement learning experiment setup (``Take the lid off saucepan"). The illustration shows train and test scenes and the task sequence. In the train scene, a green screen cloth covers the table. In the test scene, the cloth is removed, and distractors are added around the table. UR5 robots are used for this experiment, with only upper and lower wrist cameras.}
\end{figure}

\section{More Visualisations}\label{app:more-visulisations}

\begin{figure}[H]
    \centering
    \includegraphics[width=\textwidth]{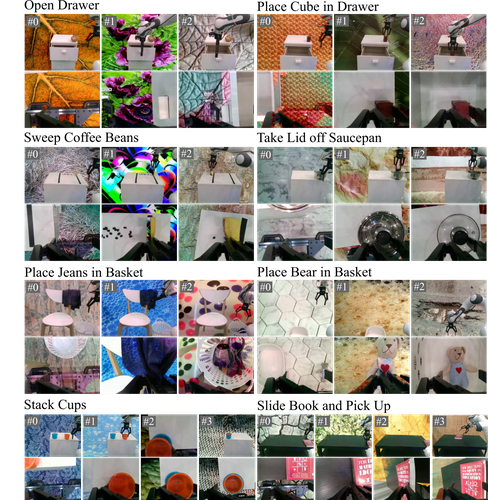}
    \caption{Visual observations of \greenrand{} with MIL textures~\citep{finn2017one}.}
    \label{fig:all_tasks_greenaug_rand}
\end{figure}

\begin{figure}[H]
    \centering
    \includegraphics[width=0.75\textwidth]{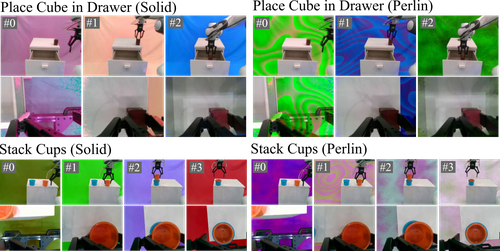}
    \caption{Visual observations of \greenrand{} with solid \& Perlin textures.}
\end{figure}

\begin{figure}[H]
    \centering
    \includegraphics[width=\textwidth]{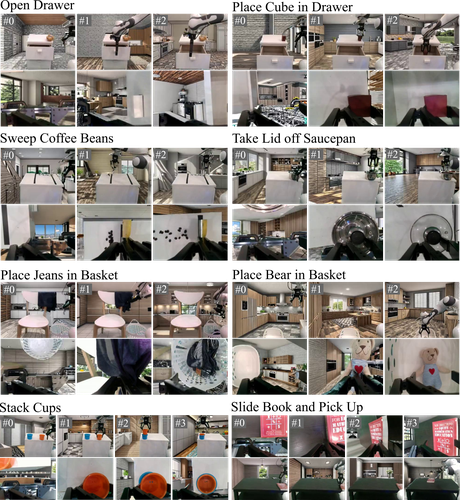}
    \caption{Visual observations of \greengen{}}
    \label{fig:all_tasks_greenaug_gen}
\end{figure}

\begin{figure}[H]
    \centering
    \includegraphics[width=0.6\textwidth]{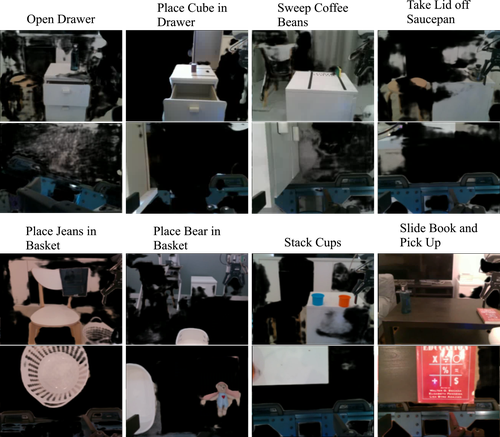}
    \caption{Visual observations of \greenmask{} during inference.}
    \label{fig:masked}
\end{figure}

\begin{figure}[H]
    \centering
    \includegraphics[width=0.75\textwidth]{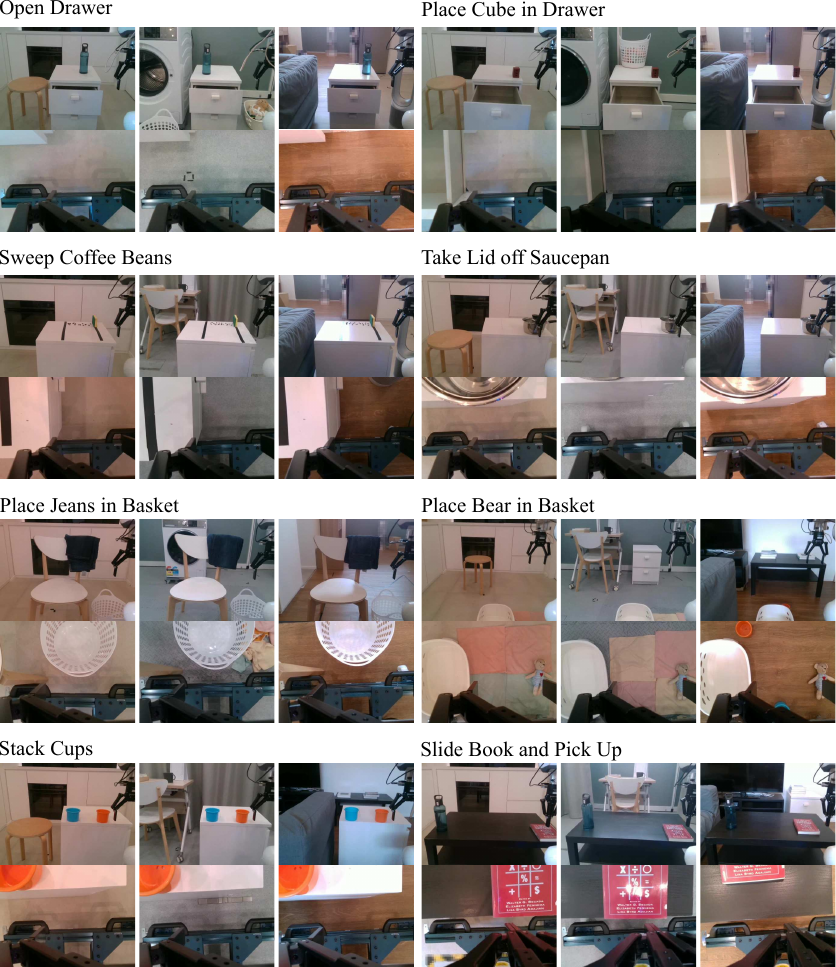}
    \caption{Visuals of three novel scenes used during evaluations for each task in respective order. These include a subset of kitchen, washing, study, and living rooms.}
    \label{fig:novel_scenes}
\end{figure}

\newpage

\section{Compute and Hyperparameter Details}

We perform the preprocessing and model training using NVIDIA L4 GPUs (24GB VRAM).

\textbf{ACT.} We use the same implementation of ACT as described in the original paper, with the following changes to hyperparameters: action chunking size is set to 20, the number of epochs is 5000, and we sample 16 transitions per epoch. Unlike the original ACT implementation, which samples one transition per episode per epoch, we sample multiple transitions. 

\begin{table}[H]
\vspace{-1em}
\caption{Pre-processing hyperparameters for each task. Chroma key parameters are represented by Key Colour ($K$) in hexadecimal colour codes, \texttt{tola} ($\alpha$), and \texttt{tolb} ($\beta$). Detection Text Prompt is used for generative augmentation. Inpaint Text Prompt is used for both generative augmentation and \greengen{} for background generation.}
\label{tab:preprocess-params}
\resizebox{\textwidth}{!}{%
\begin{tabular}{@{}cp{2cm}p{1cm}p{1cm}p{4cm}p{4cm}@{}}
\toprule
Task & Key Colour ($K$) & \texttt{tola} ($\alpha$) & \texttt{tolb} ($\beta$) & Detection Text Prompt & Inpaint Text Prompt \\ \midrule Open Drawer & \#439f82 & 30 & 35 & \texttt{drawer. robot arm. robot gripper.} & \texttt{photorealistic kitchen, study room, washing room, living room, or bedroom} \\
Place Cube in Drawer & \#25806f & 35 & 40 & \texttt{red cube. drawer. robot arm. robot gripper.} & \texttt{photorealistic kitchen, study room, washing room, living room, or bedroom} \\
\text{Sweep Coffee Beans} & \#1d6953 & 23 & 30 & \texttt{sponge. coffee beans. black tapes. drawer. robot arm. robot gripper.} & \texttt{photorealistic kitchen, study room, washing room, living room, or bedroom} \\
Take Lid off Saucepan & \#348367 & 15 & 25 & \texttt{saucepan. drawer. robot arm. robot gripper.} & \texttt{photorealistic kitchen, study room, washing room, living room, or bedroom} \\
Place Jeans in Basket & \#25806f & 30 & 40 & \texttt{jeans. chair. robot arm. robot gripper.} & \texttt{photorealistic kitchen, study room, washing room, living room, or bedroom}\\
Place Bear in Basket & \#25806f & 30 & 30 & \texttt{soft toy. basket. robot arm. robot gripper.} & \texttt{photorealistic kitchen, study room, washing room, living room, or bedroom} \\
Stack Cups & \#348367 & 15 & 25 & \texttt{blue cup. orange cup. drawer. robot arm. robot gripper.} & \texttt{photorealistic kitchen, study room, washing room, living room, or bedroom} \\ 
Slide Book and Pick Up & \#25806f & 20 & 30 & \texttt{book. table. robot arm. robot gripper.} & \texttt{photorealistic kitchen, study room, washing room, living room, or bedroom} \\
Object Generalisation & \#699230 & 30 & 20 & \texttt{green cup. drawer. robot arm. robot gripper.} & \texttt{colourful cup, bowl, cube, toy, can, bottle or general graspable object} \\ \bottomrule
\end{tabular}
}
\end{table}

\textbf{\greenmask{} U-Net.} We use the original U-Net architecture~\citep{ronneberger2015unet} (implemented by \citet{Iakubovskii:2019}) for the masking network used in \greenmask{}. The model comprises 14.3 million parameters.

\begin{table}[H]
    \centering
    \caption{Masking network hyperparameters for \greenmask{}.}
    \label{tab:mask_hyperparams}
    \begin{tabular}{@{}ll@{}}
    \toprule
    Model & Unet \\
    Encoder & ResNet18 \\
    Encoder Weights & ImageNet \\
    Epochs & 100 \\
    Batch size & 128 \\
    Image size & $224 \times 224$ \\
    Seed & 42 \\ \bottomrule
    \end{tabular}%
\end{table}

\section{Detailed Results}\label{app:detailed-results}

This section presents the full unaggregated results.

\begin{table}[H]
\centering
\caption{Full experiment results. ``Green Screen\textrightarrow Green Screen'' roughly represents the upper bound performance (in parentheses) and is not included in the average. Full unaggregated results for each task are in \cref{tab:open_drawer,tab:place_cube,tab:wipe_coffee,tab:saucepan,tab:place_jeans,tab:place_bear,tab:slide_book,tab:stack_cups}. The tables are also hyperlinked in the task text below.}
\label{tab:sum_results}
\resizebox{\textwidth}{!}{%
\begin{tabular}{@{}ccccccccc@{}}
\toprule
 &  &  & \multicolumn{6}{c}{Success Rate (\%)} \\ \cmidrule(l){4-9} 
Task & \begin{tabular}[c]{@{}c@{}}Train\\ Scene\end{tabular} & \begin{tabular}[c]{@{}c@{}}Test\\ Scene\end{tabular} & NoAug & CVAug & \begin{tabular}[c]{@{}c@{}}Generative\\ Augmentation\end{tabular} & \begin{tabular}[c]{@{}c@{}}\boldgreen\greenaug\\ random\end{tabular} & \begin{tabular}[c]{@{}c@{}}\boldgreen\greenaug\\ generative\end{tabular} & \begin{tabular}[c]{@{}c@{}}\boldgreen\greenaug\\ mask\end{tabular} \\ \midrule
\multirow{3}{*}{\hyperref[tab:open_drawer]{Open Drawer}} & Green Screen & Green Screen & (100) & (88) & (96) & (100) & (100) & (100) \\
 & Living Room & 3 Novel Scenes & 63 & 51 & 57 & - & - & - \\
 & Green Screen & 3 Novel Scenes & 55 & 79 & \textbf{96} & \textbf{96} & 87 & 79 \\ \midrule
\multirow{3}{*}{\hyperref[tab:place_cube]{Place Cube in Drawer}} & Green Screen & Green Screen & (92) & (96) & (72) & (100) & (84) & (96) \\
 & Living Room & 3 Novel Scenes & 33 & 64 & 68 & - & - & - \\
 & Green Screen & 3 Novel Scenes & 39 & 73 & 72 & \textbf{92} & 83 & 37 \\ \midrule
\multirow{3}{*}{\hyperref[tab:wipe_coffee]{Sweep Coffee Beans}} & Green Screen & Green Screen & (100) & (96) & (88) & (96) & (80) & (92) \\
 & Living Room & 3 Novel Scenes & 55 & 79 & 73 & - & - & - \\
 & Green Screen & 3 Novel Scenes & 77 & 77 & 77 & \textbf{96} & 81 & 84 \\ \midrule
\multirow{3}{*}{\hyperref[tab:saucepan]{Take Lid off Saucepan}} & Green Screen & Green Screen & (96) & (84) & (92) & (80) & (80) & (84) \\
 & Living Room & 3 Novel Scenes & 61 & 79 & 72 & - & - & - \\
 & Green Screen & 3 Novel Scenes & 73 & 83 & 83 & \textbf{88} & 73 & 71 \\ \midrule
\multirow{3}{*}{\hyperref[tab:place_jeans]{Place Jeans in Basket}} & Green Screen & Green Screen & (100) & (100) & (92) & (100) & (100) & (92) \\
 & Living Room & 3 Novel Scenes & 69 & 73 & 75 & - & - & - \\
 & Green Screen & 3 Novel Scenes & 72 & 77 & 77 & \textbf{87} & 77 & 67 \\ \midrule
\multirow{3}{*}{\hyperref[tab:place_bear]{Place Bear in Basket}} & Green Screen & Green Screen & (100) & (100) & (100) & 100 & (96) & (100) \\
 & Living Room & 3 Novel Scenes & 35 & 45 & 41 & - & - & - \\
 & Green Screen & 3 Novel Scenes & 55 & 80 & 81 & \textbf{95} & 49 & 41 \\ \midrule
\multirow{3}{*}{\hyperref[tab:stack_cups]{Stack Cups}} & Green Screen & Green Screen & (76) & (84) & (84) & (88) & (80) & (80) \\
 & Living Room & 3 Novel Scenes & 41 & 57 & 75 & - & - & - \\
 & Green Screen & 3 Novel Scenes & 57 & 61 & 80 & \textbf{81} & 72 & 55 \\ \midrule
\multirow{3}{*}{\hyperref[tab:slide_book]{Slide Book and Pick Up}} & Green Screen & Green Screen & (100) & (100) & (100) & (100) & (100) & (100) \\
 & Living Room & 3 Novel Scenes & 43 & 61 & 89 & - & - & - \\
 & Green Screen & 3 Novel Scenes & 55 & 87 & 89 & \textbf{93} & \textbf{93} & 35 \\ \midrule
\multicolumn{3}{c}{Average} & 55 & 70 & 75 & \textbf{91} & 77 & 58 \\ \bottomrule
\end{tabular}%
}
\end{table}

\begin{table}[H]
\centering
\caption{``Open Drawer" task unaggregated results.}
\label{tab:open_drawer}
\resizebox{\textwidth}{!}{%
\begin{tabular}{@{}cccccccc@{}}
\toprule
 &  & \multicolumn{6}{c}{Success Rate (\%)} \\ \cmidrule(l){3-8} 
\begin{tabular}[c]{@{}c@{}}Train\\ Scene\end{tabular} & \begin{tabular}[c]{@{}c@{}}Test\\ Scene\end{tabular} & NoAug & CVAug & \begin{tabular}[c]{@{}c@{}}Generative\\ Augmentation\end{tabular} & \begin{tabular}[c]{@{}c@{}}\boldgreen\greenaug\\ random\end{tabular} & \begin{tabular}[c]{@{}c@{}}\boldgreen\greenaug\\ generative\end{tabular} & \begin{tabular}[c]{@{}c@{}}\boldgreen\greenaug\\ mask\end{tabular} \\ \midrule
Green Screen & Green Screen & (100) & (88) & (96) & (100) & (100) & (100) \\ \midrule
\multirow{3}{*}{Living Room} & Novel Scene 1 & 60 & 48 & 68 & - & - & - \\
 & Novel Scene 2 & 68 & 64 & 64 & - & - & - \\
 & Novel Scene 3 & 60 & 40 & 40 & - & - & - \\ \midrule
\multirow{3}{*}{Green Screen} & Novel Scene 1 & 52 & 88 & 88 & 100 & 84 & 76 \\
 & Novel Scene 2 & 72 & 80 & 100 & 100 & 96 & 80 \\
 & Novel Scene 3 & 40 & 68 & 100 & 88 & 80 & 80 \\ \bottomrule
\end{tabular}%
}
\end{table}

\newpage

\begin{table}[H]
\centering
\caption{``Place Cube in Drawer" task unaggregated results..}
\label{tab:place_cube}
\resizebox{\textwidth}{!}{%
\begin{tabular}{@{}cccccccc@{}}
\toprule
 &  & \multicolumn{6}{c}{Success Rate (\%)} \\ \cmidrule(l){3-8} 
\begin{tabular}[c]{@{}c@{}}Train\\ Scene\end{tabular} & \begin{tabular}[c]{@{}c@{}}Test\\ Scene\end{tabular} & NoAug & CVAug & \begin{tabular}[c]{@{}c@{}}Generative\\ Augmentation\end{tabular} & \begin{tabular}[c]{@{}c@{}}\boldgreen\greenaug\\ random\end{tabular} & \begin{tabular}[c]{@{}c@{}}\boldgreen\greenaug\\ generative\end{tabular} & \begin{tabular}[c]{@{}c@{}}\boldgreen\greenaug\\ mask\end{tabular} \\ \midrule
Green Screen & Green Screen & (92) & (96) & (72) & (100) & (84) & (96) \\ \midrule
\multirow{3}{*}{Living Room} & Novel Scene 1 & 68 & 88 & 72 & - & - & - \\
 & Novel Scene 2 & 0 & 32 & 64 & - & - & - \\
 & Novel Scene 3 & 32 & 72 & 68 & - & - & - \\ \midrule
\multirow{3}{*}{Green Screen} & Novel Scene 1 & 36 & 92 & 76 & 92 & 92 & 48 \\
 & Novel Scene 2 & 56 & 68 & 64 & 96 & 92 & 48 \\
 & Novel Scene 3 & 24 & 60 & 76 & 88 & 64 & 16 \\ \bottomrule
\end{tabular}%
}
\end{table}

\begin{table}[H]
\centering
\caption{``Sweep Coffee Beans" task unaggregated results.}
\label{tab:wipe_coffee}
\resizebox{\textwidth}{!}{%
\begin{tabular}{@{}cccccccc@{}}
\toprule
 &  & \multicolumn{6}{c}{Success Rate (\%)} \\ \cmidrule(l){3-8} 
\begin{tabular}[c]{@{}c@{}}Train\\ Scene\end{tabular} & \begin{tabular}[c]{@{}c@{}}Test\\ Scene\end{tabular} & NoAug & CVAug & \begin{tabular}[c]{@{}c@{}}Generative\\ Augmentation\end{tabular} & \begin{tabular}[c]{@{}c@{}}\boldgreen\greenaug\\ random\end{tabular} & \begin{tabular}[c]{@{}c@{}}\boldgreen\greenaug\\ generative\end{tabular} & \begin{tabular}[c]{@{}c@{}}\boldgreen\greenaug\\ mask\end{tabular} \\ \midrule
Green Screen & Green Screen & (100) & (96) & (88) & (96) & (80) & (92) \\ \midrule
\multirow{3}{*}{Living Room} & Novel Scene 1 & 60 & 96 & 88 & - & - & - \\
 & Novel Scene 2 & 52 & 60 & 88 & - & - & - \\
 & Novel Scene 3 & 52 & 80 & 44 & - & - & - \\ \midrule
\multirow{3}{*}{Green Screen} & Novel Scene 1 & 92 & 80 & 88 & 100 & 92 & 96 \\
 & Novel Scene 2 & 76 & 80 & 88 & 96 & 80 & 76 \\
 & Novel Scene 3 & 64 & 72 & 56 & 92 & 72 & 80 \\ \bottomrule
\end{tabular}%
}
\end{table}

\begin{table}[H]
\centering
\caption{``Take Lid off Saucepan" task unaggregated results.}
\label{tab:saucepan}
\resizebox{\textwidth}{!}{%
\begin{tabular}{@{}cccccccc@{}}
\toprule
 &  & \multicolumn{6}{c}{Success Rate (\%)} \\ \cmidrule(l){3-8} 
\begin{tabular}[c]{@{}c@{}}Train\\ Scene\end{tabular} & \begin{tabular}[c]{@{}c@{}}Test\\ Scene\end{tabular} & NoAug & CVAug & \begin{tabular}[c]{@{}c@{}}Generative\\ Augmentation\end{tabular} & \begin{tabular}[c]{@{}c@{}}\boldgreen\greenaug\\ random\end{tabular} & \begin{tabular}[c]{@{}c@{}}\boldgreen\greenaug\\ generative\end{tabular} & \begin{tabular}[c]{@{}c@{}}\boldgreen\greenaug\\ mask\end{tabular} \\ \midrule
Green Screen & Green Screen & (96) & (84) & (92) & (80) & (80) & (84) \\ \midrule
\multirow{3}{*}{Living Room} & Novel Scene 1 & 64 & 76 & 76 & - & - & - \\
 & Novel Scene 2 & 68 & 84 & 68 & - & - & - \\
 & Novel Scene 3 & 52 & 76 & 72 & - & - & - \\ \midrule
\multirow{3}{*}{Green Screen} & Novel Scene 1 & 64 & 80 & 80 & 88 & 76 & 68 \\
 & Novel Scene 2 & 96 & 96 & 92 & 84 & 76 & 80 \\
 & Novel Scene 3 & 60 & 72 & 76 & 92 & 68 & 64 \\ \bottomrule
\end{tabular}%
}
\end{table}

\begin{table}[H]
\centering
\caption{``Place Jeans in Basket" task unaggregated results.}
\label{tab:place_jeans}
\resizebox{\textwidth}{!}{%
\begin{tabular}{@{}cccccccc@{}}
\toprule
 &  & \multicolumn{6}{c}{Success Rate (\%)} \\ \cmidrule(l){3-8} 
\begin{tabular}[c]{@{}c@{}}Train\\ Scene\end{tabular} & \begin{tabular}[c]{@{}c@{}}Test\\ Scene\end{tabular} & NoAug & CVAug & \begin{tabular}[c]{@{}c@{}}Generative\\ Augmentation\end{tabular} & \begin{tabular}[c]{@{}c@{}}\boldgreen\greenaug\\ random\end{tabular} & \begin{tabular}[c]{@{}c@{}}\boldgreen\greenaug\\ generative\end{tabular} & \begin{tabular}[c]{@{}c@{}}\boldgreen\greenaug\\ mask\end{tabular} \\ \midrule
Green Screen & Green Screen & (100) & (100) & (92) & (100) & (100) & (92) \\ \midrule
\multirow{3}{*}{Living Room} & Novel Scene 1 & 64 & 64 & 68 & - & - & - \\
 & Novel Scene 2 & 76 & 76 & 76 & - & - & - \\
 & Novel Scene 3 & 68 & 80 & 80 & - & - & - \\ \midrule
\multirow{3}{*}{Green Screen} & Novel Scene 1 & 68 & 76 & 76 & 84 & 72 & 64 \\
 & Novel Scene 2 & 72 & 80 & 76 & 80 & 80 & 72 \\
 & Novel Scene 3 & 76 & 76 & 80 & 96 & 80 & 64 \\ \bottomrule
\end{tabular}%
}
\end{table}

\begin{table}[H]
\centering
\caption{``Place Bear in Basket" task unaggregated results.}
\label{tab:place_bear}
\resizebox{\textwidth}{!}{%
\begin{tabular}{@{}cccccccc@{}}
\toprule
 &  & \multicolumn{6}{c}{Success Rate (\%)} \\ \cmidrule(l){3-8} 
\begin{tabular}[c]{@{}c@{}}Train\\ Scene\end{tabular} & \begin{tabular}[c]{@{}c@{}}Test\\ Scene\end{tabular} & NoAug & CVAug & \begin{tabular}[c]{@{}c@{}}Generative\\ Augmentation\end{tabular} & \begin{tabular}[c]{@{}c@{}}\boldgreen\greenaug\\ random\end{tabular} & \begin{tabular}[c]{@{}c@{}}\boldgreen\greenaug\\ generative\end{tabular} & \begin{tabular}[c]{@{}c@{}}\boldgreen\greenaug\\ mask\end{tabular} \\ \midrule
Green Screen & Green Screen & (100) & (100) & (100) & (100) & (96) & (100) \\ \midrule
\multirow{3}{*}{Living Room} & Novel Scene 1 & 32 & 24 & 32 & - & - & - \\
 & Novel Scene 2 & 36 & 24 & 12 & - & - & - \\
 & Novel Scene 3 & 36 & 88 & 80 & - & - & - \\ \midrule
\multirow{3}{*}{Green Screen} & Novel Scene 1 & 56 & 92 & 80 & 100 & 44 & 32 \\
 & Novel Scene 2 & 28 & 52 & 68 & 84 & 12 & 24 \\
 & Novel Scene 3 & 80 & 96 & 96 & 100 & 92 & 68 \\ \bottomrule
\end{tabular}%
}
\end{table}

\begin{table}[H]
\centering
\caption{``Stack Cups" task unaggregated results.}
\label{tab:stack_cups}
\resizebox{\textwidth}{!}{%
\begin{tabular}{@{}cccccccc@{}}
\toprule
 &  & \multicolumn{6}{c}{Success Rate (\%)} \\ \cmidrule(l){3-8} 
\begin{tabular}[c]{@{}c@{}}Train\\ Scene\end{tabular} & \begin{tabular}[c]{@{}c@{}}Test\\ Scene\end{tabular} & NoAug & CVAug & \begin{tabular}[c]{@{}c@{}}Generative\\ Augmentation\end{tabular} & \begin{tabular}[c]{@{}c@{}}\boldgreen\greenaug\\ random\end{tabular} & \begin{tabular}[c]{@{}c@{}}\boldgreen\greenaug\\ generative\end{tabular} & \begin{tabular}[c]{@{}c@{}}\boldgreen\greenaug\\ mask\end{tabular} \\ \midrule
Green Screen & Green Screen & (76) & (84) & (84) & (88) & (80) & (80) \\ \midrule
\multirow{3}{*}{Living Room} & Novel Scene 1 & 44 & 52 & 64 & - & - & - \\
 & Novel Scene 2 & 40 & 60 & 76 & - & - & - \\
 & Novel Scene 3 & 40 & 60 & 84 & - & - & - \\ \midrule
\multirow{3}{*}{Green Screen} & Novel Scene 1 &  56 & 44 & 76 & 72 & 60 & 24 \\
 & Novel Scene 2 & 64 & 80 & 88 & 84 & 72 & 80 \\
 & Novel Scene 3 & 52 & 60 & 76 & 88 & 84 & 60 \\ \bottomrule
\end{tabular}%
}
\end{table}

\begin{table}[H]
\centering
\caption{``Slide Book and Pick Up" task unaggregated results.}
\label{tab:slide_book}
\resizebox{\textwidth}{!}{%
\begin{tabular}{@{}cccccccc@{}}
\toprule
 &  & \multicolumn{6}{c}{Success Rate (\%)} \\ \cmidrule(l){3-8} 
\begin{tabular}[c]{@{}c@{}}Train\\ Scene\end{tabular} & \begin{tabular}[c]{@{}c@{}}Test\\ Scene\end{tabular} & NoAug & CVAug & \begin{tabular}[c]{@{}c@{}}Generative\\ Augmentation\end{tabular} & \begin{tabular}[c]{@{}c@{}}\boldgreen\greenaug\\ random\end{tabular} & \begin{tabular}[c]{@{}c@{}}\boldgreen\greenaug\\ generative\end{tabular} & \begin{tabular}[c]{@{}c@{}}\boldgreen\greenaug\\ mask\end{tabular} \\ \midrule
Green Screen & Green Screen & (100) & (100) & (100) & (100) & (100) & (100) \\ \midrule
\multirow{3}{*}{Living Room} & Novel Scene 1 & 68 & 84 & 88 & - & - & - \\
 & Novel Scene 2 & 28 & 48 & 84 & - & - & -  \\
 & Novel Scene 3 & 32 & 52 & 96 & - & - & - \\ \midrule
\multirow{3}{*}{Green Screen} & Novel Scene 1 & 44 & 96 & 92 & 96 & 96 & 32 \\
 & Novel Scene 2 & 48 & 72 & 84 & 88 & 88 & 8 \\
 & Novel Scene 3 & 72 & 92 & 92 & 96 & 96 & 64 \\ \bottomrule
\end{tabular}%
}
\end{table}

\newpage

\begin{table}[H]
\centering
\caption{Texture randomness unaggregated results (\greenrand{}). ``Green Screen\textrightarrow Green Screen'' roughly represents the upper bound performance (in parentheses) and is not included in the average.}
\resizebox{\textwidth}{!}{%
\begin{tabular}{@{}ccccccc@{}}
\toprule
\multicolumn{3}{c}{} & \multicolumn{4}{c}{Success Rate (\%)} \\ \cmidrule(l){4-7} 
Task & \begin{tabular}[c]{@{}c@{}}Train\\ Scene\end{tabular} & \begin{tabular}[c]{@{}c@{}}Test\\ Scene\end{tabular} & None & Solid Textures & Perlin Textures & MIL Textures \\ \midrule
\multirow{4}{*}{\begin{tabular}[c]{@{}c@{}}Place Cube\\ in Drawer\end{tabular}} & Green Screen & Green Screen & (92) & (100) & (100) & (100) \\ \cmidrule(l){2-7}
 & \multirow{3}{*}{Green Screen} & Novel Scene 1 & 36 & 68 & 64 & 92 \\
 & & Novel Scene 2 & 56 & 36 & 56 & 96 \\
 & & Novel Scene 3 & 24 & 68 & 68 & 88 \\ \midrule
\multirow{4}{*}{Stack Cups} & Green Screen & Green Screen & (76) & (76) & (80) & (88) \\ \cmidrule(l){2-7}
 & \multirow{3}{*}{Green Screen} & Novel Scene 1 & 56 & 76 & 68 & 72 \\ 
 & & Novel Scene 2 & 64 & 68 & 60 & 84 \\
 & & Novel Scene 3 & 52 & 76 & 80 & 88 \\ \midrule
\multicolumn{3}{c}{Average} & 48 & 65 & 66 & \textbf{87} \\ \bottomrule
\end{tabular}%
}
\end{table}

\begin{table}[H]
\centering
\caption{Green screen coverage unaggregated results (\greenrand{}). ``Green Screen\textrightarrow Green Screen'' roughly represents the upper bound performance (in parentheses) and is not included in the average.}
\resizebox{0.80\textwidth}{!}{%
\begin{tabular}{@{}cccccccc@{}}
\toprule
\multicolumn{3}{c}{} & \multicolumn{5}{c}{Success Rate (\%)} \\ \cmidrule(l){4-8} 
Task & \begin{tabular}[c]{@{}c@{}}Train\\ Scene\end{tabular} & \begin{tabular}[c]{@{}c@{}}Test\\ Scene\end{tabular} & 0\% & 25\% & 50\% & 75\% & 100\% \\ \midrule
\multirow{2}{*}{\begin{tabular}[c]{@{}c@{}}Place Cube\\ in Drawer\end{tabular}} & Green Screen & Green Screen & (92) & (100) & (100) & (100) & (100) \\ \cmidrule(l){2-8}
 & \multirow{3}{*}{Green Screen} & Novel Scene 1 & 36 & 80 & 76 & 80 & 92 \\ 
& & Novel Scene 2 & 56 & 64 & 68 & 72 & 96 \\ 
& & Novel Scene 3 & 24 & 88 & 84 & 88 & 88 \\ \midrule
\multirow{2}{*}{Stack Cups} & Green Screen & Green Screen & (76) & (76) & (80) & (76) & (88) \\ \cmidrule(l){2-8}
 & \multirow{3}{*}{Green Screen} & Novel Scene 1 & 56 & 64 & 76 & 80 & 72 \\ 
& & Novel Scene 2 & 64 & 68 & 60 & 68 & 84 \\ 
& & Novel Scene 3 & 52 & 72 & 76 & 76 & 88 \\ \midrule
\multicolumn{3}{c}{Average} & 48 & 73 & 73 & 77 & \textbf{87} \\ \bottomrule
\end{tabular}%
}
\end{table}

\begin{table}[H]
\centering
\caption{Object generalisation unaggregated results. Data is collected on the green cup, and policies are then trained and evaluated on various objects (illustrated in \cref{fig:obj_gen}).}
\begin{tabular}{@{}cccc@{}}
\toprule
 & \multicolumn{3}{c}{Success Rate (\%)} \\ \cmidrule(l){2-4} 
Object Type & NoAug & \greenrand & \greengen \\ \midrule
Green Cup & \textbf{96} & 88 & 80 \\
Blue Cup & \textbf{96} & 80 & 80 \\
Orange Cup & \textbf{92} & 80 & 80 \\
Red Cube & 0 & 44 & \textbf{60} \\
Green Cube & 0 & 20 & \textbf{44} \\
Soda Can & \textbf{40} & 28 & 36 \\
Soya Can & 36 & \textbf{64} & 44 \\
Soft Toy & 0 & \textbf{84} & 72 \\ \midrule
Average & 45 & 61 & \textbf{62} \\ \bottomrule
\end{tabular}%
\end{table}%

\begin{figure}[H]
  \begin{subfigure}{0.5\textwidth}
    \centering
    \includegraphics[width=0.95\linewidth]{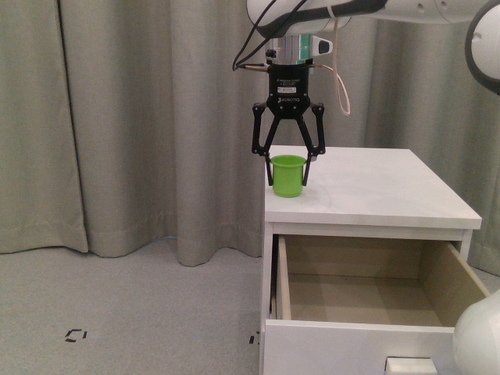}
    \caption{Demonstrations collected using the green cup.}
    \end{subfigure}
  \begin{subfigure}{0.5\textwidth}
    \centering
    \includegraphics[width=0.95\linewidth]{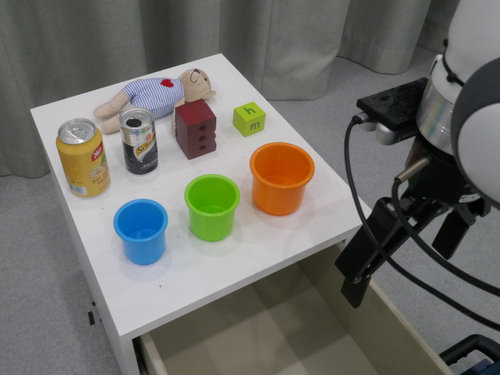}
    \caption{Test objects: cups, cans, cubes and toy.}
    \end{subfigure}
    \caption{Object generalisation across object category. Policy trained on green cup data using \greenrand{} and tested on different objects.}
    \label{fig:obj_gen}
\end{figure}

\section{Code Snippet for CVAug}

The following code snippet outlines the data augmentation pipeline for CVAug. The pipeline begins by applying a random photometric distortion to the image. Next, it performs a random shift by cropping the image to its original size and adding a 4-pixel padding on each side. Each augmentation is wrapped in RandomApply, meaning each has a 50\% chance of being applied. As a result, there's a 50\% chance of applying the photometric distortion, a 50\% chance of performing the random shift, and a 25\% chance of both augmentations being applied simultaneously.

\begin{verbatim}
import torchvision.transforms.v2 as TV

original_image_size = (240, 320)

TV.Compose(
    [
        TV.RandomApply([T.RandomPhotometricDistort()]),
        TV.RandomApply(
            [TV.RandomCrop(original_image_size, padding=4)]
        ),  # random shift
    ]
)
\end{verbatim}

\section{Additional Limitations and Future Works}

\textbf{Exploration of better chroma key algorithms.} The chroma key algorithm used in this paper~\citep{chromakey} is a basic one that performs reasonably well, but it does not produce perfect masks. Some parameter tuning for $K$, $\alpha$, and $\beta$ is still necessary. Despite these imperfections, we demonstrate that \greenaug{} still significantly outperforms the baselines. In the film industry, extensive manual post-processing is often required to achieve perfect masks~\citep{wright2017digital}. Future research could explore more advanced chroma key algorithms that provide superior green screen masks~\citep{aksoy2016interactive,li2021automatic,jin2022automatic,smirnov2023magenta}. This could potentially enhance the performance of \greenmask{}, which relies heavily on green screen mask as ground truth for training.

\textbf{Pose generalisation.} A major ongoing challenge in robot learning is generalising to 6D poses not present in the training dataset. Current robot learning policies, especially imitation learning-based ones often fail when objects are relocated to different positions within 3D space.

\textbf{Application to methods with 3D observations.} Currently, \greenaug{} has only been tested on RGB-based robot learning policies. Recent advances in next-best-pose-based agents~\citep{james2022coarse,shridhar2023peract,ma2024hierarchical} have demonstrated that by aligning the observation space with action space, we can obtain strong generalisation in robot learning policies. As a general plug-and-play method, \greenaug{} could potentially further improve the scene generalisation of the next-best-pose agents, which we leave for future study.

\end{document}